\documentclass[runningheads]{llncs}
\usepackage{graphicx}
\usepackage{hyperref}

\usepackage{times}
\usepackage{array}
\usepackage{latexsym}
\usepackage{url}
\usepackage{multirow}
\usepackage[utf8]{inputenc}
\usepackage{xcolor}
\usepackage{booktabs}

\def\tmtodo#1{{\color{purple}TM: #1}}
\def\dmtodo#1{{\color{red}DM: #1}}
\def\rrtodo#1{{\color{blue}RR: #1}}
\def\tmtodo#1{}
\def\dmtodo#1{}
\def\rrtodo#1{}

\begin{document}
%
\title{Measuring Memorization Effect in Word-Level Neural Networks Probing\thanks{This work has been supported by grant 18-02196S of the Czech Science Foundation.
It has been using language resources and tools developed, stored and distributed by the LINDAT/CLARIAH-CZ project of the Ministry of Education, Youth and Sports of the Czech Republic (project LM2018101)}}
%
%
\author{Rudolf Rosa\orcidID{0000-0003-4908-6127} \and
Tomáš Musil\orcidID{0000-0002-4013-560X} \and\\
David Mareček\orcidID{0000-0001-5327-488X}}
\authorrunning{R. Rosa et al.}
%
\institute{Charles University, Faculty of Mathematics and Physics, Institute of Formal and Applied Linguistics, Malostranské náměstí 25, 118 00 Praha, Czechia\\
\email{\{rosa,musil,marecek\}@ufal.mff.cuni.cz}\\
\url{https://ufal.mff.cuni.cz/}}
\maketitle              
\begin{abstract}
%
%

%
Multiple studies have probed representations emerging in neural networks trained for end-to-end NLP tasks and examined what word-level linguistic information may be encoded in the representations.
In classical probing, a classifier is trained on the representations to extract the target linguistic information.
However, there is a threat of the classifier simply memorizing the linguistic labels for individual words, instead of extracting the linguistic abstractions from the representations, thus reporting false positive results.
While considerable efforts have been made to minimize the memorization problem, the task of actually measuring the amount of memorization happening in the classifier has been understudied so far.
In our work, we propose a simple general method for measuring the memorization effect, based on a symmetric selection of comparable sets of test words seen versus unseen in training.
Our method can be used to explicitly quantify the amount of memorization happening in a probing setup, so that an adequate setup can be chosen and the results of the probing can be interpreted with a reliability estimate.
We exemplify this by showcasing our method on a case study of probing for part of speech in a trained neural machine translation encoder.

%

\end{abstract}
%
%
%

\section{Introduction}
\label{sec:intro}



In recent years, there has been a considerable amount of research into linguistic abstractions emerging in neural networks trained for various natural language processing (NLP) tasks.
It has been found that, to some degree, neural networks often capture abstractions which seem to correspond to classical linguistic notions known from the linguistic studies of morphology, syntax or semantics, even if they were not explicitly trained to do so.
The common hypothesis is that modern neural networks are sufficiently powerful to unravel many linguistic properties and regularities of language, and that they do so if this is useful for solving the task for which they are trained.


In this work, we focus on the subfield of identifying word-level linguistic abstractions, such as part-of-speech (POS) labels, in word-level representations, such as static or contextual word embeddings.

The usual method of assessing the amount to which linguistic abstractions are captured by a neural network is to use \textit{probing}, which we review in Section~\ref{sec:relwork:probing}.
In word-level probing, we take representations of words from a trained neural network (such as word embeddings or hidden states from an encoder) and train a classifier to predict linguistic labels (such as POS) from the representations corresponding to the words, using linguistically annotated data (such as a tagged corpus).
The common assumption is that if the classifier learns to predict the linguistic labels with a high accuracy, it is an indication that the neural word representations contain a latent abstraction similar to the linguistic notion (e.g.\ that contextual word embeddings encode POS of the words).

\subsection{The Memorization Problem}

A major threat associated with the probing approach is that of \textit{memorization}.
As the probing classifier learns to assign labels to words, it can succeed in two ways.
Either, it learns to extract an abstraction from the word representation which corresponds to the label to assign; this is the intended case, which we refer to as \textit{generalization}.
Or, it simply memorizes the label associated with each word; we refer to this as \textit{memorization}.
If memorization occurs, the result of the probing can be misinterpreted as the representations capturing some linguistic abstractions, while the actual underlying mechanism is that the representations simply capture the word identity. The probing classifier thus only learns to extract the word identity from the representation and memorizes the label for the word.\footnote{%
Unlike static word embeddings, contextual representations of the same word in different sentences are different, which makes memorization harder, but not impossible:
the identity of the word is still strongly encoded in the contextual representation and can be extracted from it, especially when a stronger classifier is used.}
A crucial problem is that, without taking additional measures, there is no way of distinguishing the true positive result from the false positive result.




With context-independent word representations (static word embeddings), it is of course possible to avoid the problem by splitting the vocabulary into two disjoint sets of words, training the classifier on a train set and testing it on a test set.
However, for contextual representations, this cannot be done easily, as the representations need to be computed for whole sentences, not for individual words, and the train and test sets thus need to be composed of full sentences, which unavoidably have a large word overlap.
While we might evaluate the probe only on test set words unseen in the training data, these are not representative of the language, as such a set of test words will be biased towards low-frequency words.
We argue that we rather need to evaluate on the full test set while measuring and minimizing the memorization effect.

\subsection{Measuring Memorization}


In this paper, we suggest a general method of measuring the amount of memorization occurring in word-level probing of neural network representations,
based on comparing the probing classifier accuracy on sets of \textit{seen} and \textit{unseen} words.
Although a standard test set contains both words seen and unseen in training data, the seen words tend to be frequent while the unseen ones are typically rare words; we thus regard an approach of comparing accuracies on these sets of words as inadequate and uninformative.
Instead, we propose a method which samples the seen and unseen words in a symmetric way to ensure their comparability.

We do not present a new method for probing itself; our method is designed to complement existing probing approaches by explicitly measuring their reliability with respect to the memorization problem.
This can help the researcher to select an adequate probing setup by providing means for quantifying the magnitude of the memorization problem, allowing for a trustworthy interpretation of the probing results.




As a case study, we apply our method to measure the amount of memorization in probing for POS in word representations from a neural machine translation system.

\section{Related Work}

A comprehensive survey of word embeddings evaluation methods was compiled by Bakarov \cite{bakarov2018survey}. An overview can also be found in the survey of methodology for analysis of deep learning models for NLP by Belinkov and Glass~\cite{belinkov2019analysis}.
Another overview \cite{faruqui2016problems} mentions “[n]o standardized splits \& overfitting” as one of the problems of evaluating word embeddings with similarity tasks.

\label{sec:relwork:probing}

There are various strategies when it comes to the train/dev/test splitting in probing.

When it is possible to predict the probed property from the word type itself, the vocabulary may be split into train/test sets. This strategy is used e.g. in \cite{qian2016investigating,musil_examining_2019} to evaluate POS tag and other morphological features prediction.

Some works split the dataset into train/dev/test sets, without regard to the same words occuring in both. These include  
 predicting syntactic and semantic labels (including POS) from hidden states on sentences  \cite{shi-etal-2016-string,belinkov-etal-2017-neural,belinkov-etal-2017-evaluating,dalvi-etal-2017-understanding,linzen:2016:TACL}
 or treebanks \cite{blevins-etal-2018-deep,kohn2015s}.

Bisazza and Tump \cite{bisazza-tump-2018-lazy} address the problem with the overlap. They observe that even a dummy random feature can be predicted with high accuracy when the same words occur both in the train and the test data. They extract one vector per token from the NMT encoder. They randomly split the vocabulary into two parts and use one to filter the training data and the other to filter the test data. They repeat the experiments several times and report mean accuracies.

Another approach to evaluating words in context of sentences is presented by \cite{conneau2018you}. They propose the word content task that tests whether it is possible to recover information about the original words in the sentence from its embedding. They pick 1000 mid-frequency words from the source corpus vocabulary and sample equal numbers  of  sentences  that  contain  one  and  only one of these words. The words can then be partitioned into train and test sets without the risk of their overlapping.

\label{sec:relwork:memorization}

The ability of deep neural networks to memorize is a challenge for the theory of deep learning \cite{arpit_closer_2017}. It also has implications for the applications of neural networks, because it may be problematic if a portion of the training data can be reconstructed from the trained model \cite{carlini_secret_2019}.

In connection with probing neural networks, memorization was addressed by Hewitt and Liang \cite{hewitt_designing_2019}, who propose control tasks to complement the linguistics tasks. A control task associates word types with random labels. If the classifier performs well on the control task, this means that it is able to memorize the training set.
However, the data distribution affects the generalization ability of deep neural networks and they tend to learn  simple patterns when possible \cite{krueger_deep_2017}. 
Our approach differs from \cite{hewitt_designing_2019} by using the original data to measure the memorization effect, evading the problem created by altering the distribution in a control task.

\section{Method}
\label{sec:method}

In the usual probing approach, we operate with two sets of sentences, a training set and a test set, both labelled with the word-level labels corresponding to the linguistic abstraction for which we are probing the neural word representations (e.g.\ POS).
The training set is used to train a probing classifier to predict the labels from the word representations. The classifier is then evaluated on the test set, and its accuracy, compared to a baseline, is used to estimate to what extent the given linguistic abstraction is encoded in the word representations.

The goal of our method is to measure to what extent the probing classifier only memorizes word identities instead of measuring the generalization captured by the word representations.
The main idea
is to compare the probing classifier accuracies on words that are part of the training data (\textit{seen} words) and on words that are not (\textit{unseen} words), while keeping the sets of seen and unseen words otherwise comparable (as discussed in Section~\ref{sec:intro}), which we ensure by a symmetric way of creating these sets.

We propose the following approach:
\begin{enumerate}
    \item Randomly split the training set into two halves, which we will refer to as \textit{seen sentences} and \textit{unseen sentences}.
    \item Train the probing classifier only on the seen sentences.
    \item Apply the probing classifier to the test set.
    \item Define the set of \textit{seen words} as words that are contained in the seen sentences but not in the unseen sentences.
    \item Define the set of \textit{unseen words} as words that are contained in the unseen sentences but not in the seen sentences.
    \item Evaluate the accuracy of the probing classifier separately on seen words and on unseen words, ignoring words that are neither seen nor unseen.\footnote{Note that words which occur in both seen and unseen sentences are neither seen words nor unseen words. We also need to remove words that are part of the development set if one is used for training the probing classifier. Technically, words that do not appear in the test set can also be removed from the sets of seen and unseen words as they do not influence the results.}
\end{enumerate}



Using this approach, we can now quantify the magnitude of the memorization effect occurring in the probing setup as the difference between the classifier accuracy on seen and on unseen words.
If the memorization problem is not present, these accuracies should be identical, as the classifier only extracts the linguistic abstraction from the representation, regardless of the word identity; in this case, the classifier accuracy reliably measures the amount of linguistic information encoded by the representation.
On the other hand, a higher accuracy on seen words than on unseen words signalizes that the classifier memorized some of the seen words' identities to some extent, instead of extracting the linguistic abstractions from them.

To stabilize the evaluation, we propose to sample the seen and unseen sentences and train the classifier multiple times, and to compute the microaverage accuracy.




We define our method as operating on words and word representations, as this makes the subsequent word-level probing straightforward. 
Our method is in principle applicable even for setups using subwords.
However, in such cases, it is up to the researcher to decide whether for the given language and setup, subword-level memorization is a problem or not, as our method only deals with word-level memorization.

\subsection{Which Words Are Selected for Evaluation?}
\label{sec:discussion}

It is important to note that the distribution of words selected for evaluation by our method is strongly biased towards lower-frequency words.
Very frequent words are never selected for evaluation, and medium-frequency words are rarely selected, as they always or nearly always appear in both seen and unseen sentences, and our method is thus unable to measure the memorization effect for such words.

Specifically, the probability $P_{sel}(w)$ of a word $w$ being selected as \textit{unseen} (or \textit{seen}) follows a hypergeometric distribution:
$
P_{sel}(w) \sim \mathrm{Hypergeometric}\left(|S|, \frac{|S|}{2}, |S_w|\right)
$, 
where $S$ is the set of training sentences, out of which its subset $S_w$ contains the word $w$. 
For most words,\footnote{For frequent words, the actual probability is even lower than the (already negligible) approximated value; for words that appear in more than half of the training sentences, the probability is $0$. The probability is also technically $0$ for words that do not appear in the test set.}
it is similar to the binomial distribution $\mathrm{Bi}(|S_w|, 0.5)$, and $P_{sel}(w)$ is thus inversely exponentially proportional to $|S_w|$:
$
P_{sel}(w) \approx \left(\frac{1}{2}\right)^{|S_w|}
$.


We believe that for \textbf{very frequent words} (especially function words such as common prepositions, pronouns, determiners and punctuation), avoiding memorization is hard -- a set of sentences constructed not to contain a given word from this class would typically not be very representative of the language.
Moreover, the probed neural network is typically not very likely to meaningfully abstract over such words, as it is usually more economical for the network to simply memorize the most frequent words and treat them as special cases.\footnote{Which they often are, as frequent words tend to behave irregularly in language \cite[p.~116]{davies2008handbook}.}\fnmsep\footnote{Arguably, it is sane to memorize very frequent words rather than abstracting over them. Nevertheless, we should be able to measure this reliably, not mistaking one for the other.}

For \textbf{medium-frequency words}, such as common nouns and verbs, we see their underrepresentation as a shortcoming of our method which we intend to focus on in future work.
We specifically plan to further investigate the approach of Bisazza and Tump \cite{bisazza-tump-2018-lazy}, reviewed in Section~\ref{sec:relwork:probing}, who train the probing classifier on representations of only some words in the training sentences and regard the other words as \textit{unseen}. We appreciate the approach, but we believe that it must be analyzed to what degree it may be influenced by the contextual representations of the \textit{seen} words containing information about surrounding words regarded as \textit{unseen}.\footnote{In their method, \textit{unseen} words are part of the training sentences and can thus influence the contextual representations of the \textit{seen} words which are used for training the probing classifier, whereas in our method, the training sentences do not contain the \textit{unseen} words at all.}

Our method mostly focuses on \textbf{lower-frequency words}, which we believe to be reasonable, as the lower the frequency of the word, the stronger is the network forced to abstract over the word.
We are thus mostly interested in such words in probing, as if the network captures the abstractions that we are probing for, they should be most prominent in representations of lower-frequency words.

Still, we also omit \textbf{very rare words}, which either do not appear in the test sentences or in the training sentences (or, obviously, in none of those).
For these words, the memorization effect is very unlikely to occur.

\section{Case Study}
\label{sec:study}

\begin{table}[tb]
\begin{tabular}[b]{@{}r@{\hskip 1 em}lc@{\hskip .25 em}r@{\hskip 1 em}rc}
\toprule
 &
\multicolumn{3}{c}{\textbf{Accuracy}\hskip 1 em\hbox{}} &
\multicolumn{2}{c}{\bf Stand. dev.} \\ \cmidrule(r{1em}){2-4}\cmidrule{5-6}

\textbf{Train sent.} &
\textbf{seen} & \textbf{unseen} & \textbf{diff} &
\textbf{seen} & \textbf{unseen}\\
\addlinespace
\midrule
\multicolumn{6}{c}{Encoder output states, linear classifier}\\
\midrule
50     & 90.5 & 87.3 & 3.3  & 3.4 & 5.6 \\
100    & 89.1 & 86.8 & 2.3  & 1.8 & 2.0 \\
500    & 93.9 & 92.8 & 1.1  & 0.9 & 1.1 \\
1,000  & 94.7 & 93.9 & 0.8  & 0.9 & 0.8 \\
5,000  & 95.5 & 94.9 & 0.7  & 0.5 & 0.6 \\
10,000 & 95.7 & 95.5 & 0.2  & 0.8 & 0.8 \\
30,000 & 95.8 & 95.9 & 0.0  & 0.4 & 0.4 \\
\addlinespace
\midrule
\multicolumn{6}{c}{Encoder output states, MLP}\\
\midrule
50     & 97.7 & 93.3 & 4.4  & 1.5 & 3.2 \\
100    & 96.2 & 93.6 & 2.7  & 1.0 & 1.4 \\
500    & 97.2 & 94.5 & 2.7  & 0.3 & 0.9 \\
1,000  & 96.8 & 94.9 & 1.9  & 0.7 & 0.7 \\
5,000  & 97.6 & 95.7 & 1.9  & 0.4 & 0.5 \\
10,000 & 98.0 & 96.2 & 1.8  & 0.7 & 0.7 \\
30,000 & 97.7 & 96.1 & 1.6  & 0.6 & 0.7 \\
\bottomrule
\end{tabular}\hfill\begin{tabular}[b]{@{}r@{\hskip 1 em}lc@{\hskip .25 em}r@{\hskip 1 em}rc}
\toprule
 &
\multicolumn{3}{c}{\textbf{Accuracy}\hskip 1 em\hbox{}} &
\multicolumn{2}{c}{\bf Stand. dev.} \\ \cmidrule(r{1em}){2-4}\cmidrule{5-6}

\textbf{Train sent.} &
\textbf{seen} & \textbf{unseen} & \textbf{diff} &
\textbf{seen} & \textbf{unseen}\\
\addlinespace
\midrule
\multicolumn{6}{c}{Encoder word embeddings, linear classifier}\\
\midrule
50     & 98.5 & 74.3 & 24.1 & 0.9 & 7.6 \\
100    & 97.0 & 78.0 & 19.0 & 0.8 & 2.3 \\
500    & 97.6 & 80.5 & 17.1 & 0.7 & 3.2 \\
1,000  & 97.0 & 82.8 & 14.2 & 1.0 & 1.5 \\
5,000  & 96.2 & 84.7 & 11.4 & 0.5 & 1.7 \\
10,000 & 95.2 & 85.3 & 10.0 & 0.8 & 1.0 \\
30,000 & 93.5 & 88.0 & 5.4  & 0.6 & 1.3 \\
\addlinespace
\midrule
\multicolumn{6}{c}{Encoder word embeddings, MLP}\\
\midrule
50     & 98.5 & 76.6 & 21.8 & 0.9 & 6.9 \\
100    & 97.0 & 81.4 & 15.6 & 0.7 & 3.0 \\
500    & 97.8 & 87.4 & 10.3 & 0.4 & 1.9 \\
1,000  & 97.7 & 89.8 & 7.9  & 0.5 & 1.4 \\
5,000  & 98.4 & 92.7 & 5.6  & 0.2 & 1.0 \\
10,000 & 98.7 & 93.5 & 5.2  & 0.2 & 1.0 \\
30,000 & 98.4 & 94.2 & 4.1  & 0.6 & 1.2 \\
\bottomrule
\end{tabular}
\caption{Case study evaluation on POS prediction, varying the number of training sentences, the probed representations, and the probing classifier. The difference between the accuracy of the probe on seen versus unseen words represents the magnitude of the memorization problem. Micro-average over 10 repetitions, in percentage points, with standard deviations.}
\label{tab:results}
\end{table}

As a case study, we apply our method to probing representations from a neural machine translation model for POS.
We study the memorization phenomenon along three dimensions, varying the train set size, the contextuality of the representation (static word embeddings versus encoder output states), and the power of the probing classifier, using either a linear classifier or a multi-layer perceptron classifier (MLP).


We analyze a Transformer model \cite{vaswani2017attention} implemented within the Neural Monkey framework\footnote{\url{https://github.com/ufal/neuralmonkey}}~\cite{helcl_neural_2018}, trained for the task of machine translation from Czech to English on the CzEng dataset\footnote{\url{http://ufal.mff.cuni.cz/czeng}}~\cite{czeng16:2016}.
The setup is based on \cite{libovickysolving}, with the exception of splitting the sentences into words instead of subwords, as explained in Section~\ref{sec:method}; we use a vocabulary of 25,000 words that are most frequent in the parallel training data.

We probe the source word embeddings and source encoder output states for Universal POS with a linear classifier (softmax) or a MLP with one hidden layer of dimension 512, using the Universal Dependencies 1.4 version of the Czech Prague Dependency Treebank \cite{ud14}.
We use the first 500 sentences from the treebank training data as tuning data for the probing classifier, the rest of the training data is used to create the seen and unseen sentence sets, using either the full data or subsampling smaller subsets.
The probing classifier is then evaluated using the development part of the treebank using token-based evaluation.
For each setup, we repeat the experiment 10 times with different samples of the seen and unseen sentences and report micro-average results.

By comparing the accuracies of the probing classifier on seen and unseen words in Table~\ref{tab:results}, we can see that the memorization problem is clearly most pronounced with static word embeddings, where the magnitude of the effect (the difference in the accuracies) ranges from 4 points for the full training set up to 24 points for a training set of 50 sentences, while for the contextual representations, the effect does not surpass 5 points.
The memorization effect is more pronounced with the stronger classifier, and disappears only with the linear classifier applied to contextual representations when trained with the largest train set.

\section{Conclusion}

We presented a method for measuring the memorization effect in word-level probing of neural representations of words, based on a comparison of the accuracy of the probing classifier on symmetrically sampled comparable sets of \textit{seen} and \textit{unseen} words.
As we showed in a case study on probing for POS, our method can measure the magnitude of the memorization problem and can thus serve as a means for selecting an appropriate probing setup, as well as for estimating the reliability of the findings of the probing experiment with respect to the threat of mistaking memorization for generalization.

In future, we intend to tackle the shortcoming of our method of underrepresenting medium-frequency words.
We also plan to apply the method to a wider range of word-based probing tasks, as well as to measure the memorization effect for existing previous probing works and reassess results reported by their authors from this perspective.



\bibliographystyle{splncs04}
\bibliography{references}
\end{document}